\documentclass{bmvc2k}

\def \hfillx {\hspace*{-\textwidth} \hfill}

\usepackage{siunitx}
\usepackage[flushleft]{threeparttable}
\usepackage{booktabs}

\usepackage{graphicx}

\title{A Fine-to-Coarse Convolutional Neural Network for 3D Human Action Recognition}

\addauthor{Thao Minh Le}{thao@ks.cs.titech.ac.jp}{1}
\addauthor{Nakamasa Inoue}{inoue@ks.cs.titech.ac.jp}{1}
\addauthor{Koichi Shinoda}{shinoda@c.titech.ac.jp}{1}

\addinstitution{
 Department of Computer Science\\
 Tokyo Institute of Technology\\
 Tokyo, Japan
}

\runninghead{Thao Le et al.}{A Fine-to-Coarse CNN for 3-D Human Action Recognition}

\begin{document}

\maketitle

\begin{abstract}
This paper presents a new framework for human action recognition from a 3D skeleton sequence. Previous studies do not fully utilize the temporal relationships between video segments in a human action. Some studies successfully used very deep Convolutional Neural Network (CNN) models but often suffer from the data insufficiency problem. In this study, we first segment a skeleton sequence into distinct temporal segments in order to exploit the correlations between them. The temporal and spatial features of a skeleton sequence are then extracted simultaneously by utilizing a fine-to-coarse (F2C) CNN architecture optimized for human skeleton sequences. We evaluate our proposed method on NTU RGB+D and SBU Kinect Interaction dataset. It achieves 79.6\% and 84.6\% of accuracies on NTU RGB+D with cross-object and cross-view protocol, respectively, which are almost identical with the state-of-the-art performance. In addition, our method significantly improves the accuracy of the actions in two-person interactions.
\end{abstract}

\section{Introduction}
	\label{sec:intro}

\begin{figure}[t!]
	\centering
	\includegraphics[width=\columnwidth, clip]{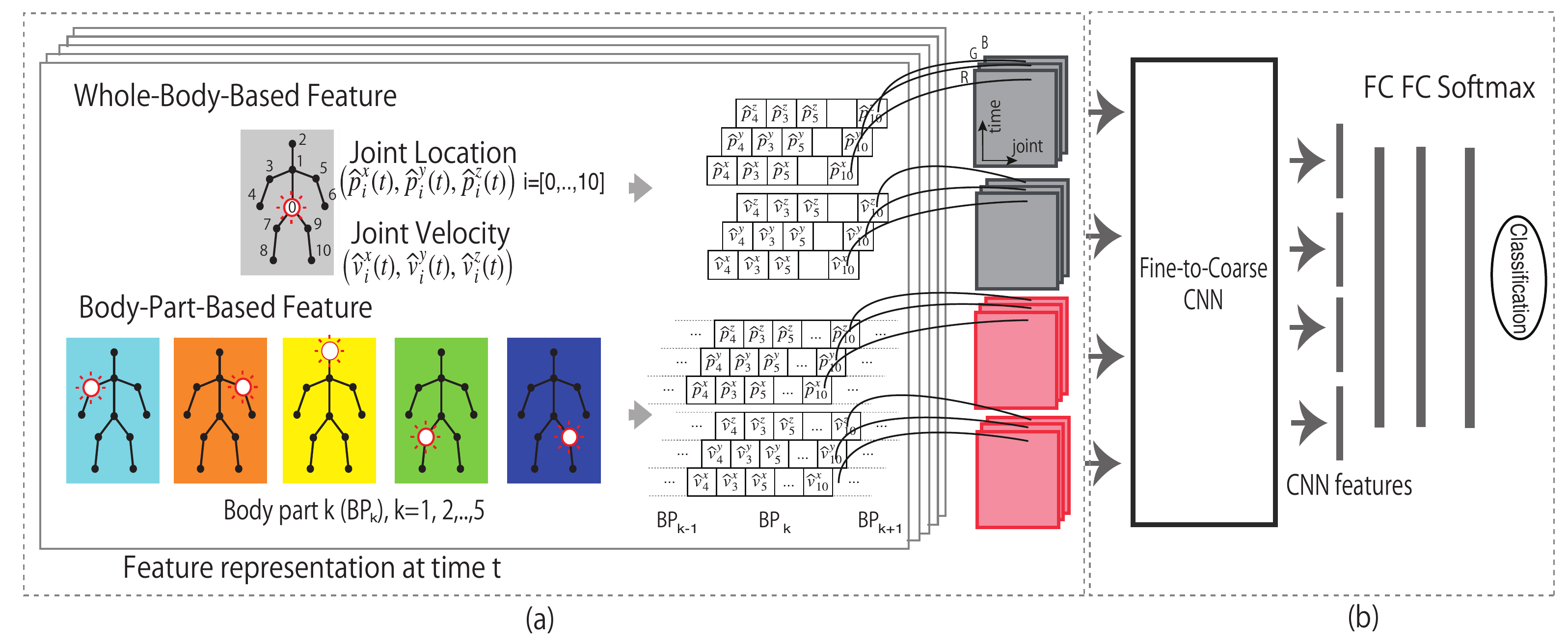}
	\caption{Overview of The Proposed Method. It consists of two parts: (a) feature representation and (b) high-level feature learning with a F2C CNN-based network architecture. A skeleton from a video input sequence is represented by whole-body-based features (WB) and body-part-based features (BP). These features are then transformed into a skeleton image that contains both the spatial structure information of a human body as well as the temporal sequence feature of a human action. The skeleton images are then fed into an F2C convolutional neural network for high-level feature learning. Finally, CNN features are concatenated before being passed to two subsequent fully connected layers, and a soft-max layer for final classification.}
	\label{fig:network_arch}
\end{figure}
In the past few years, human action recognition has become an intensive area of research, as a result of the dramatic growth of societal applications for a number of areas including security surveillance systems, human-computer-interaction-based games, and healthcare industry. The conventional approach based on RGB data was not robust against intra-class variations and illumination variations. With the advancement of 3D sensing technologies, in particular, affordable RGB-D cameras such as Microsoft Kinect, these problems have been remedied to some extent. Human action recognition studies utilizing 3D skeleton data have drawn a great deal of attention \cite{han2017space, presti20163d}.

Human action recognition based on 3D skeleton data is a time series problem, and accordingly, a great body of previous studies have focused on extracting motion patterns from a skeleton sequence. Earlier methods utilized hand-crafted features for representing the intra-frame relationships through the skeleton sequences \cite{yang2014effective, wang2012mining}. Some studies utilized the deep learning, end-to-end learning based on Recurrent Neural Networks (RNNs) with Long Short-Term Memory (LSTM) has been utilized to learn the temporal dynamics \cite{du2015hierarchical, song2017end, zhu2016co, liu2016spatio, shahroudy2016ntu, liu2017skeleton}. Recent studies have shown the superiority of Convolutional Neural Networks (CNNs) over RNN with LSTM for this task \cite{ke2017skeletonnet, liu2017enhanced, ke2017new, liu20173d}. Most of the CNN-based studies encoded the trajectories of human joints in an image space representing the spatio-temporal information of the skeleton data. The encoded feature is then fed into a deep CNN pre-trained on large scale image datasets, for example, ImageNet \cite{ILSVRC15}, under the notion of transfer learning \cite{pan2010survey}. This CNN-based method is, however, weak in handling long temporal sequences. And thus, it usually fails to distinguish actions with similar distance variations but with different durations, such as ``handshaking'' and ``giving something to other persons''. 

Motivated by the success of the generative model for CAPTCHA images  \cite{george2017generative}, we believe 3D human action recognition systems can also benefit from a specific network structure for this application domain. The first step is to segment a given skeleton sequence into different temporal segments. Here, we assume that temporal features of different time-steps have different correlations. We further utilize a tailor-made F2C CNN-based network architecture to model high-level features. By utilizing both the temporal relationships between temporal segments and spatial connectivities among human body parts, our method is expected to have a superior performance to the naive deep CNN networks. To the best of our knowledge, this is the first attempt to use F2C network for 3D human action recognition.

The paper is organized as follows. In Section \ref{sec:rl_studies}, we discuss the related studies. In Section \ref{sec:approach}, we explain our proposed network architecture in detail. We then show the experimental results to justify our motivations in Section \ref{sec:exps}. Finally, we conclude our study in Section \ref{sec:conclusion}.

\section{Related studies}
	\label{sec:rl_studies}
% Related studies
Deep learning techniques drew a great attention in the field of 3D human action recognition. Especially, the end-to-end network architectures can discriminate actions from raw skeleton data without any handcrafted features. Zhu et al.  \cite{zhu2016co} adopted three LSTM layers to exploit the co-occurrence features of skeleton joints at different layers. Du et al. \cite{du2015hierarchical} proposed a hierarchical RNN to exploit the spatio-temporal feature of a skeleton sequence. They divided skeleton joints into five subsets corresponding to five body parts before independently feeding them into five bidirectional recurrent neural networks for local feature extraction. The relationships between body parts were then modeled in later layers by hierarchically fusing them together. LSTMs were deliberately used in the last layer to tackle the vanishing problem of a vanilla RNN.

The use of deep learning techniques for this area of research was exploded when NTU RGB+D dataset \cite{shahroudy2016ntu} was released. Shahroudy et al. \cite{shahroudy2016ntu} introduced a part-aware LSTM to learn the long-term dynamics of a long skeleton sequence from multimodal inputs extracted from human body parts. Liu et al. \cite{liu2016spatio}, on the other hand, employed a spatio-temporal LSTM (ST-LSTM) to handle both the spatial dependency and the temporal dependency. ST-LSTM is also enhanced with a tree-structure based traversal method for transmitting input data of each frame into the network. 
In addition, this method used a trust gate mechanism to exclude noisy data from the input. Zhang et al. \cite{zhang2017view} proposed a view adaptation scheme for 3D skeleton data and further integrated it into an end-to-end LSTM network for sequential data modeling and feature extraction. 

CNNs are powerful for the task of object detection from images. Transfer learning techniques enable them to perform well even with a limited number of data samples \cite{wagner2013learning, long2015learning}. Motivated by this, Ke et al. \cite{ke2017skeletonnet} was the first to apply transfer learning for 3D human action recognition. They used a VGG model \cite{Chatfield14} pre-trained with ImageNet to extract high-level features from cosine distance features between joint vectors and their normalized magnitude. Ke et.al 2017b \cite{ke2017new} further transformed the cylindrical coordinates of an original skeleton sequence into three clips of gray-scale images. The clips are then processed by pre-trained VGG19 model \cite{simonyan2014very} to extract image features. Multi-task learning was also proposed by \cite{ke2017new} for the final classification, which achieved the state-of-the-art performance on NTU RGB+D dataset.

Our study addresses two problems of the previous studies: (1) the loss of temporal information of a skeleton sequence during training and, (2) the need for a specific CNN structure for skeleton data. We believe that a very deep CNN model such as VGG \cite{simonyan2014very}, AlexNet \cite{krizhevsky2012imagenet} or ResNet \cite{he2016deep} are overqualified for such sparse data as human skeleton. Moreover, the available skeleton datasets are relatively small compared to image datasets. Thus, we believe a network architecture which is able to leverage the geometric dependencies of human joints is promising for solving this issue. 
    
\section{Fine-to-Coarse CNN for 3D Human Action Recognition}
	\label{sec:approach}
This section presents our proposed method for 3D skeleton-based action recognition which exploits the geometric dependency of human body parts and the temporal relationship in a time sequence of skeletons (Figure \ref{fig:network_arch}).  It consists of two phases: feature representation and high-level feature learning with a F2C network architecture.

\subsection{Feature Representation}
\begin{figure}[t!]
	\centering
	\includegraphics[width=\columnwidth, clip]{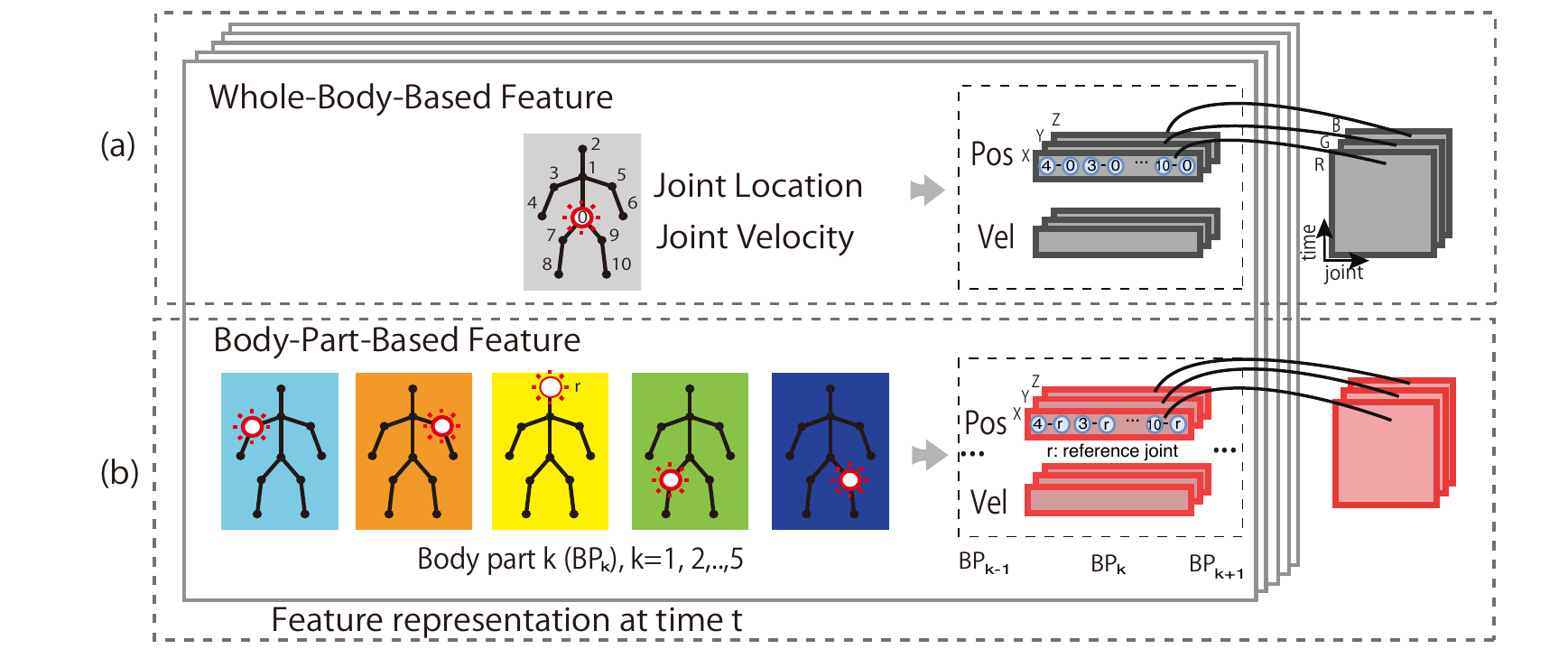}
	\caption{Feature Generation. Figure (a) illustrates the procedure of generating WB features obtained by transforming the joint positions in the camera coordinate system to the hip-based coordinate system. In Figure (b), we arrange BP features side by side to obtain one unique feature 2D array before projecting the coordinates in Euclidean space into RGB image space using a linear transformation and further up-scaling by using cubic interpolation transformation.}
	\label{fig:wb_bp}
\end{figure}
We encode the geometry of human body originally given in an image space into local coordinate systems to extract the relative geometric relationships among human joints in a video frame. We select six joints in a human skeleton as reference joints in order to generate whole-body-based (WB) features and body-part-based (BP) features. The hip joint is chosen as the origin of the coordinate system presenting the WB features, while the other reference joints, namely the head, the left shoulder, the right shoulder, the left hip, and the right hip, are selected exactly the same as \cite{ke2017skeletonnet} to represent the BP features. The WB features represent the motions of human joints around the base of the spine,  while the BP features represent the variation of appearance and deformation of the human pose when viewed from different body parts. We believe that the combined use of WB and BP is robust against coordinate transformations.

 Different from the other studies using BP features \cite{shahroudy2016ntu, liu2016spatio, ke2017skeletonnet}, we extract a velocity together with a joint position from each joint of the raw skeleton. The velocity represents the variations over the time and has been widely employed in many previous studies, mostly in the handcrafted-feature-based approaches \cite{zanfir2013moving, kerola2016graph, zhang2017geometric}. It is robust against the speed changes; and accordingly, is effective to discriminate actions with similar distance variations but with different speeds, such as punching and pushing.
 
 In the $t$-th frame of sequence of skeletons with $n$ joints, the 3D position of the $i$-th joint is depicted as:
 \begin{eqnarray}
 p_i(t)=[p_i^x(t), p_i^y(t), p_i^z(t)]^\top.
 \end{eqnarray}
 The relative inter-joint positions are highly discriminative for human actions \cite{luo2013group}. The relative position of joint $i$ at time $t$ is described as:
 \begin{eqnarray}
 \hat{p}_i(t)=p_i(t)-p_{\textrm{ref}}(t),
 \end{eqnarray}
 where $p_{\textrm{ref}}(t)$ depicts the position of a selected reference joint. The velocity feature $\hat{v}_i(t)$ at time frame $t$ is defined as the first derivatives of the relative position feature $\hat{p}_i(t)$. Zanfir et al.  \cite{zanfir2013moving} showed that it is effective to compute the derivatives of human instantaneous pose which is represented by joints' location at a given time frame $t$ over a time segment. The velocity feature, therefore, is formulated as:
 \begin{eqnarray}
 \hat{v}_i(t) \approx \hat{p}_i(t+1)-\hat{p}_i(t-1).
 \end{eqnarray}
 
\subsubsection{Whole-body-based Feature}

As mentioned above, we choose the hip joint as the reference joint in order to represent WB features (See figure \ref{fig:wb_bp}(a)). In addition, we follow the limb normalization procedure \cite{zanfir2013moving} to reduce the problem caused by the variations in human body size among human subjects. We first compute the average bone lengths of each two connected joints over the training dataset, and then use them to normalize each human subject's bones. To put it differently, we stretch each bone of a certain human subject with a normalized length while keeping the joint angle between any two bones unchanged.

In order to extract the spatial features of a human skeleton at time $t$ over the set of joints, we first define a spatial configuration of a joint chain. We believe that the order of joints greatly affects the learning ability of 2D CNN since the joints in adjacent body parts share more spatial relations than a random pair of joints. For example, in most actions, the joints of the right arm are more correlated to those of the left arm than those of the left leg are. With this intention, we concatenate joints in the following order: left arm, right arm, torso, left leg, right leg. Note that the torso in the context of this paper includes the head joint of the human skeleton. Let $T$ be the number of frames in a given skeleton sequence. In the next step, we compute each feature of skeleton data over $T$ frames and stack them as a feature row. Consequently, we obtain the WB features of two 2D arrays; each corresponds to the joint location and velocity. Finally, we project these 2D array features into RGB image space using a linear transformation. In particular, each of three components $(x, y, z)$ of each skeleton joint is represented as one of the three corresponding components $(R, G, B)$ of a pixel in a color image; by normalizing the $(x, y, z)$ values to the range 0 to 255. The two sets of color images are further up-scaled by using a cubic spline interpolation. Cubic spline interpolation is a commonly used technique in image processing to minimize the interpolation error \cite{hou1978cubic}. We call these two RGB images as skeleton images.
\subsubsection{Body-part-based Feature} 
In order to represent the BP features, we choose five joints corresponding to five human body parts as the reference joints: the head, the left shoulder, the right shoulder, the left hip, and the right hip, as in \cite{ke2017skeletonnet}. They are relatively stable in most actions. Provided that, we calculate joint position features and velocity features for each reference joint in the above order dependently. 
As a result, with each skeleton at time $t$, we obtain five feature vectors of a joint location and five vectors of a velocity corresponding to five distinct reference joints. We then place all BP features side by side to produce one unique row feature and place them along the temporal axis to obtain a 2D array feature. Finally, we apply a linear transformation to represent these array features as RGB images and further up-scale them by using a cubic spline interpolation. After all, we obtain two BP-base skeleton images; one corresponding to the joint location and the other to the velocity from each skeleton sequence. The whole process is illustrated in Figure \ref{fig:wb_bp}(b).

\subsection{Fine-to-Coarse Network Architecture}
\begin{figure}[t!]
	\centering
	\includegraphics[width=9.5cm, clip]{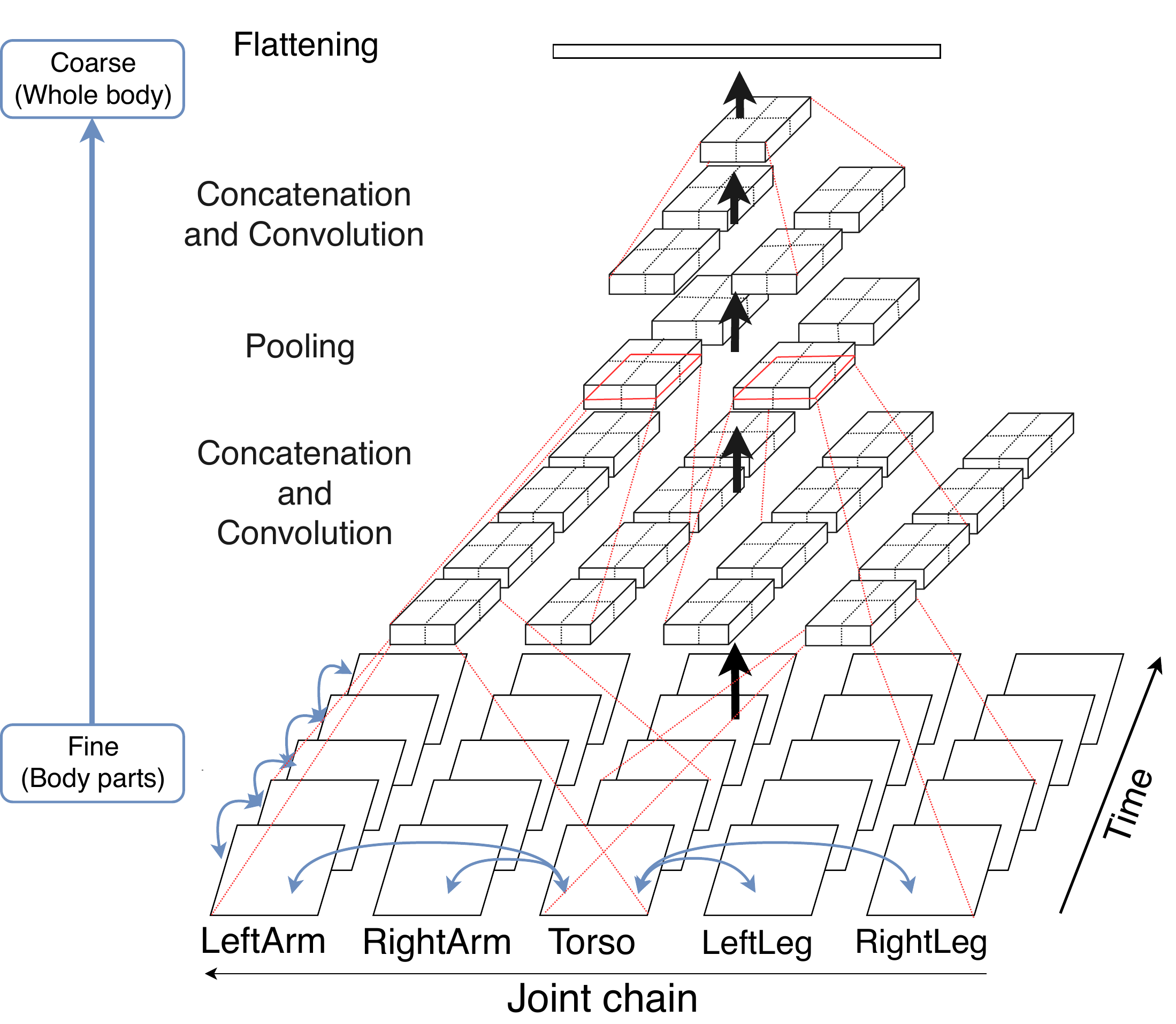}
	\caption{Proposed Fine-to-Coarse Network Architecture. Blue arrows show pair slices which are concatenated along each dimension before passing to a convolutional block.}
	\label{fig:hier_arch}
\end{figure}
In this section, we explain the detail of our proposed F2C network architecture for high-level feature learning. Figure \ref{fig:hier_arch} illustrates our network structure in three dimensions.

Our F2C network takes three color channels of skeleton images generated from the feature representation phase as inputs. Accordingly, the input of our F2C network consists of two dimensions: the spatial dimension which describes the geometric dependencies of human joints along the joint chain, and the temporal dimension of the time-feature representation over $T$ frames of a skeleton sequence. Let $m$ be the number of segments along the temporal axis , $n$ is the number of body parts ($n=5$), each image skeleton is considered as a set of $m \times n$ slices (Figure \ref{fig:hier_arch}). Assume $T_\textrm{seg}$ ($T$=m$\times$$T_\textrm{seg}$) is the number of frames in one temporal segment, $l_\textrm{bp}$ is the dimension of one body part along the spatial dimension, each input slide has size of $l_\textrm{bp}\times T_\textrm{seg}$. In the next step, we simultaneously concatenate the slices over both the spatial axis and temporal axis. In other words, regarding the spatial dimension, we first concatenate each body part which belongs to human limbs (arms and legs) with the torso, while concatenating two consecutive temporal segments together. Each concatenated 2D array feature is further passed through a convolutional layer and a max pooling layer. The same fusion procedure is applied before passing the next convolutional layer. In short, our F2C network composes of three layer-concatenation steps, and three convolutional blocks accordingly. In the last step, the extracted image features are flattened to obtain an output of 1D array feature.

Both WB-based skeleton images and BP-based skeleton images are fed into the proposed F2C network in the same way. While it is conceivable for feeding BP features into our network for high-level feature learning, we believe WB features also benefit from going through the network since the spatial dimension of WB features, which are formed by the pre-defined joint chain, includes the intrinsic relationships between body parts. 

Our network can be viewed as a procedure to eliminate unwanted connections between layers from the conventional CNN. We believe traditional CNN models include some redundant connections for capturing human-body-geometric features. Many actions only require the movement of the upper body ({\textit{e}.\textit{g}.} hand waving, clapping) or the lower body ({\textit{e}.\textit{g}.} sitting, kicking), while the other requires the movements of the whole body ({\textit{e}.\textit{g}.} moving towards another person, pick up something). For this reason, the bottom layers in our proposed method can discriminate ``fine''actions which require the movements of some certain body parts, while the top layers are discriminative for ``coarse'' actions using the movements of the whole body.

\begin{figure}[t!]
	\centering
	\includegraphics[width=7.5cm, clip]{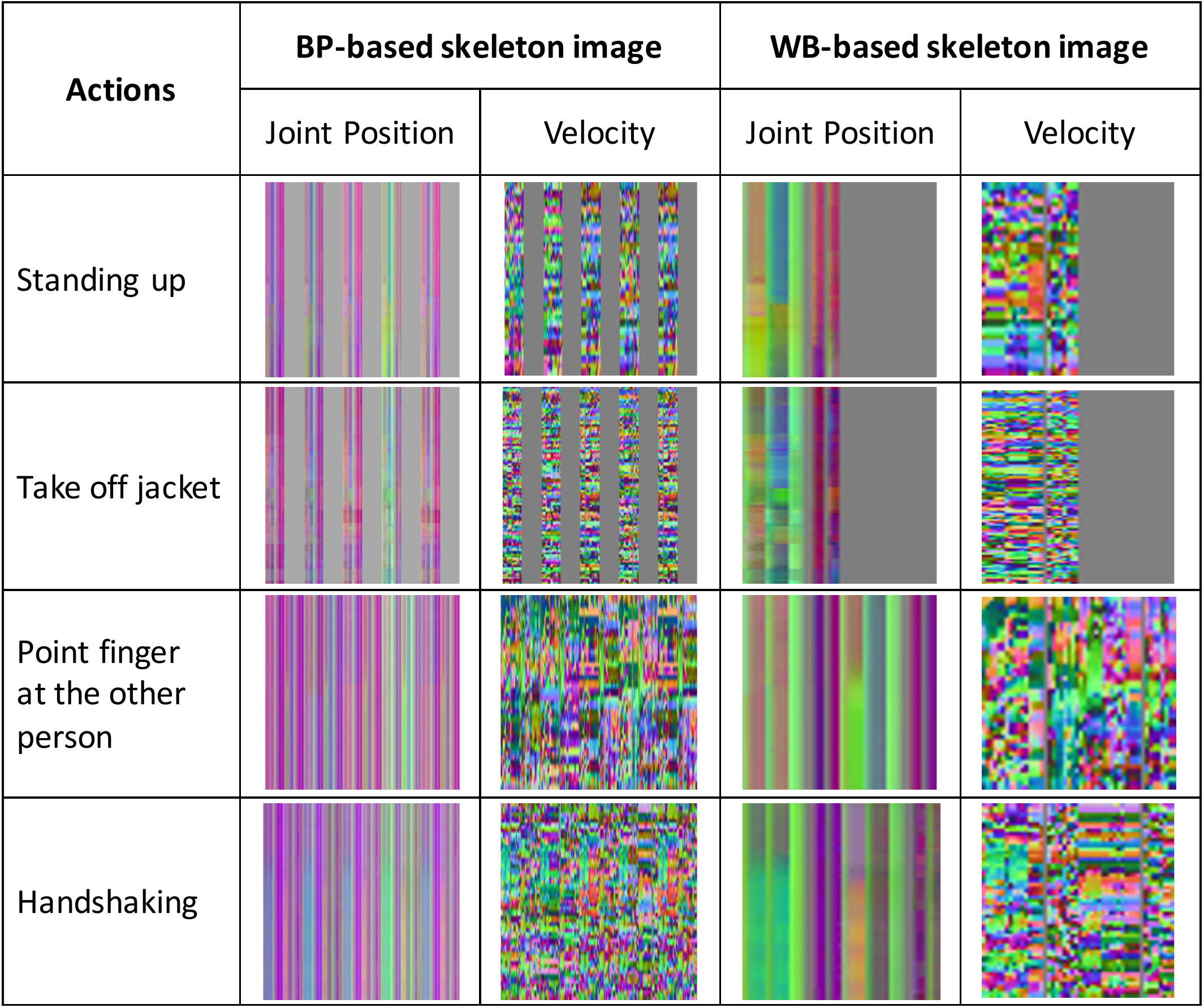}
	\caption{Examples of Generated Skeleton Images. "Standing up" and "take off jacket" present single actions while "point finger at the other person" and "handshaking" are two-person interation actions.}
	\label{fig:skeleton_images}
\end{figure}
    
\section{Experiments and discussion}
	\label{sec:exps}
\subsection{Datasets and Experimental Conditions}
\hspace*{1.25em}
We conduct experiments on two skeleton benchmark datasets publicly available: NTU RGB+D \cite{shahroudy2016ntu} and SBU Kinect Interaction Dataset \cite{yun2012two}. As the method proposed by \cite{ke2017skeletonnet} is relatively related to this paper, we employ their method as our baseline. We also compare our proposed method with other state-of-the-art methods reported on the same datasets.

\textbf{NTU RGB+D Dataset}
is the largest skeleton-based human action dataset for the time being with 56,880 sequences. The skeleton data were collected by utilizing Microsoft Kinect v2 sensors. Each skeleton contains 25 human joints. In this dataset, there are 60 distinct action classes of three human-action groups: daily actions, health-related actions, and two-person interactive actions. All the actions are performed by 40 distinct subjects. The actions are recorded simultaneously by three camera sensors located at different angles: \ang{-45}, \ang{0}, \ang{45}. This dataset is challenging due to the large variations of viewpoints and sequence lengths.
In our experiments, we use the two standard evaluation protocols proposed by the original study \cite{shahroudy2016ntu}, namely, cross-subject (CS) and cross-view (CV).%For the cross-subject evaluation, the skeleton sequences of 20 subjects are used for training and that of the 20 remaining subjects are used for testing. In particular, there are 40,320 and 16,560 samples are used for training and testing, respectively. In case of cross-view evaluation, the sequences captured by camera 2 and 3 are used for training, while the remaining is used for testing. In short, the training and testing set, in this case, have 37,920 and 18,960 samples subsequently. 

%\textbf{Experiments with small amount of data}

%This experiment aims to justify the domain-specific network architecture's generation capability as well as the reference capability (In consideration whether we need this experiment since I do not get better result than SkeletonNet in this experiment).

\begin{table}[t]
	\footnotesize 
	%\small
	\begin{minipage}{0.4\textwidth}
			\centering %\begin{center}
		\caption{Network Configuration}
		\label{tab:config}
		\begin{tablenotes}
			\scriptsize
			\item \hspace{0.3cm} conv3-64: 3$\times$3 convolution, 64 filters\\
		\end{tablenotes}
		\begin{tabular}{|c|}
			\hline
			Input of 224$\times$224 RGB image\\
			\hline
			35 input slices of 32$\times$44\\
			\hline
			24 input slices of 64$\times$88\\
			\hline
			conv3-64\\
			maxpool\\
			conv3-64\\
			maxpool\\
			\hline
			10 fused feature slices of 32$\times$44\\
			\hline
			conv3-128\\
			maxpool\\
			conv3-128\\
			maxpool\\
			\hline
			4 fused feature slices of 16$\times$22\\
			\hline
			conv3-256\\
			maxpool \\
			conv3-256\\
			maxpool \\
			\hline
			output 4$\times$5,120 \\
			\hline
		\end{tabular}
	\end{minipage}
	\hfillx
	\begin{minipage}{0.6\textwidth}
		\centering %\begin{center}
		\caption{Classification Performance on NTU RGB+D Dataset}
		\label{tab:table1}
		\begin{tabular}{|l|r|r|}
			\hline
			\multicolumn{1}{|c|}{Methods} & \multicolumn{1}{c|}{CS}& \multicolumn{1}{c|}{CV}\\
			\hline
			Lie Group \cite{vemulapalli2014human} & 50.1 & 52.8\\
			Part-aware LSTM \cite{shahroudy2016ntu} & 62.9 & 70.3 \\
			ST-LSTM + Trust Gate \cite{liu2016spatio} & 69.2 & 77.7 \\
			Temporal Perceptive Network \cite{hutemporal} & 75.3 & 84.0 \\
			Context-aware  attention LSTM \cite{liu2018skeleton} & 76.1 & 84.0 \\
			Enhanced skeleton visualization \cite{liu2017enhanced} & 76.0 & 82.6 \\
			Temporal CNNs\cite{kim2017interpretable} & 74.3 & 83.1 \\
			Clips+CNN+Concatenation \cite{ke2017new} & 77.1 & 81.1 \\
			Clips+CNN+MTLN \cite{ke2017new} & \textbf{79.6} & 84.8 \\
			VA-LSTM \cite{zhang2017view} & 79.4 & \textbf{87.6} \\
			\hline
			SkeletonNet \cite{ke2017skeletonnet} & 75.9 & 81.2 \\
			\hline
			(WB + BP) + VGG & 68.1 & 72.4\\ 
			BP + F2C network & 78.2 & 81.9 \\
			(WB + BP) w/o velocity + F2C network & 76.6 & 81.7 \\ 
			\textbf{F2CSkeleton (Proposed)}& \textbf{79.6} & 84.6\\ 
			\hline
		\end{tabular}
	\end{minipage}
\end{table}
\textbf{SBU Kinect Interaction Dataset}
is another skeleton-based dataset collected using the Microsoft Kinect sensor. There are 282 skeleton sequences divided into 21 subsets, which are collected from eight different types of two-person interactions including approaching, departing, pushing, kicking, punching, exchanging objects, hugging, and shaking hands. Each skeleton contains 15 joints. There are seven subjects who performed the actions in the same laboratory environment. We also augment data as in \cite{ke2017skeletonnet} before doing five-fold cross-validation. Each skeleton image is first resized to 250 $\times$ 250 and then is randomly cropped into 20 sub-images with the size of 224 $\times$224. Eventually, we obtain a dataset of 11,280 samples.

\textbf{Implementation Details}
The proposed model was implemented using Keras~\footnote{\url{https://github.com/keras-team/keras}} with TensorFlow backend. For a fair comparison with the previous studies, transfer learning is applied in order to improve the classification performance. To be more specific, our proposed F2C network architecture is first trained with ImageNet with the input image dimension is set to 224$\times$224. The pre-trained weights are then applied to all experiments.

Regarding input skeletons at each time step, we consider up to two distinct human subjects at once. This means that in case of two-person interactions, joint position features of the two subjects at a certain frame are read simultaneously and place side by side. In the case of single actions, we use zero matrices in the presentation of the second subject. Figure \ref{fig:skeleton_images} shows some examples of skeleton images generated from NTU RGB+D dataset.

For NTU RGB+D dataset, we first remove 302 missing skeletons reported by \cite{shahroudy2016ntu}. 20\% of training samples are used as a validation set. The first fully connected layer has 256 hidden units, while the output layer has the same size as the number of actions in the datasets. The network is trained using Adam for stochastic optimization \cite{kingma2014adam}. The learning rate is set to 0.001 and exponentially decayed over 25 epochs. We use a batch size of 32. The same experimental settings are applied to all the experiments. 

We set the number of temporal segments to seven, because it shows the best performance on NTU RGB+D dataset. Considering body part features have different contributions to an action, we do not share weights between input slices during training. This might increase the number of parameters but gain better generalization ability of the network. Table \ref{tab:config} shows the detail of our network configuration. 
\subsection{Experimental Results}
\begin{table}[t]
	%\small
	\footnotesize 
	\begin{minipage}{0.55\textwidth}
	\centering
	\caption{Classification Performance with Two-person Interactions, RGB+D Dataset, CV Protocol}
	\label{tab:table3}
	\hspace*{-1cm}
		\begin{tabular}{|l@{\hspace{0.5\tabcolsep}}|r@{\hspace{0.5\tabcolsep}}|@{\hspace{0.5\tabcolsep}} r|r@{\hspace{0.5\tabcolsep}}|@{\hspace{0.5\tabcolsep}} r|}
		\hline
		\multicolumn{1}{|c|}{Actions} & \multicolumn{2}{c|}{SkeletonNet} & \multicolumn{2}{c|}{F2CSkeleton} \\
		{}& Prec. & Rec. & Prec. & Rec. \\
		\hline
		Punching/slapping & 59.2 & 56.0 & \textbf{80.6} & \textbf{82.2}\\
		Kicking& 46.8 & 64.9 & \textbf{90.4} & \textbf{91.3}\\
		Pushing & 69.7 & 72.2 & \textbf{88.0} & \textbf{86.1}\\
		Pat on back& 54.7 & 46.2 & \textbf{82.8} & \textbf{80.7}\\
		Point finger& 42.8 & 72.8 & \textbf{88.3} & \textbf{91.1}\\
		Hugging& 77.6 & 83.5 & \textbf{92.9} & \textbf{83.8}\\
		Giving something & 72.5 & 72.5 & \textbf{88.7} & \textbf{91.8}\\
		Touch other's pocket & 66.9 & 50.6 & \textbf{90.9} & \textbf{95.3}\\
		Handshaking& 83.1 & 82.6 & \textbf{95.8} & \textbf{94.9}\\
		Walking towards& 66.2 & 82.3 & \textbf{96.9} & \textbf{97.8}\\
		Walking apart& 61.8 & \textbf{78.5} & \textbf{76.2} & 77.7\\
		\hline
		\end{tabular}
	\hspace*{-1cm}
	\begin{tablenotes}
		\footnotesize
		\item * Prec.: Precision \hspace{0.5cm}  Rec.: Recall\\
	\end{tablenotes}
	\end{minipage}
	\hfillx
	\begin{minipage}{0.43\textwidth}
		\caption{Classification Performance on SBU Dataset}
		\label{tab:table2}
		\begin{tabular}{|l@{\hspace{0.5\tabcolsep}}|r|}
			\hline
			\multicolumn{1}{|c|}{Methods} & Acc.\\
			\hline
			Deep LSTM+Co-occurence \cite{zhu2016co} & 90.4\\
			ST-LSTM+Trust Gate \cite{liu2016spatio} & 93.3\\
			SkeletonNet \cite{ke2017skeletonnet} & 93.5 \\
			Clips+CNN+Concatenation \cite{ke2017new} & 92.9 \\
			Clips+CNN+MTLN \cite{ke2017new} & 93.6 \\
			Context-aware  attention LSTM \cite{liu2018skeleton} & 94.9 \\
			VA-LSTM \cite{zhang2017view} & 97.2\\
			\hline
			\textbf{F2CSkeleton (Proposed)} & \textbf{99.1}\\ 
			\hline
		\end{tabular}
	\end{minipage}
\end{table}
\hspace{5.5mm}\textbf{NTU RGB+D Dataset}
We compare the performance of our method with the previous studies in Table \ref{tab:table1}. The classified accuracy is chosen as the evaluation metric. 

\textit{(WB + BP) + VGG} In this experiment, we use VGG16 pre-trained on ImageNet dataset instead of our F2C network. This experiment examines the significance of the proposed F2C network for high-level feature learning against the conventional deep CNN models. 

\textit{BP + F2C network} In this experiment, we only adopt the skeleton images generated by BP features to feed into the proposed F2C network architecture. This aims to justify the contribution of WB features going through our F2C network.

\textit{(WB + BP) w/o velocity + F2C network} In this experiment, only joint position features are put into the proposed F2C network architecture for the purpose of examining the importance of incorporating velocity feature to the final classification performance.

\textit{WB + BP + F2C network (F2CSkeleton)} This is our proposed method.

As shown in Table \ref{tab:table1}, our proposed method outperforms results reported by \cite{vemulapalli2014human, shahroudy2016ntu,liu2016spatio,hutemporal,liu2018skeleton,liu2017enhanced,ke2017skeletonnet} with the same testing condition. In particular, we gain over 3.0\% improvement from our baseline \cite{ke2017skeletonnet} on both CS and CV testing protocols. Similarly, our method is around 2.5\% better than the method with feature concatenation \cite{ke2017new}. However, \cite{ke2017new} with Multi-Task Learning Network (MTLN) obtained a slightly better performance than our method with the CV protocol. The learning paradigm MTLN works as a hierarchical method to effectively learn the intrinsic correlations between multiple related tasks \cite{zhang2014regularization}, thus, outperforms a mere concatenation. We believe our method also can benefit from MTLN. We will include this as a part of our future work to improve our network. It is also to note that while our method with the CS protocol outperforms \cite{zhang2017view}, they achieved a better performance of 3\% when coming to the CV protocol handled by a view adaption scheme in a multiple-view environment.

Table \ref{tab:table1} also shows that our F2C network performs significantly better than VGG16. In particular, our F2C network improves the accuracy from 68.1\% to 79.6\% with the CS protocol and from 72.4\% to 84.6\% with the CV protocol. The incorporation of velocity improves the performance about 3.0 points in both testing protocols. Besides, the use of WB and BP features in combination improves the accuracies from 78.2\% to 79.6\% and 81.9\% to 84.6\% with the CS and CV protocol, respectively.

Our method outperforms SkeletonNet on all the two-person interactions. Table \ref{tab:table3} shows our classification performance with the CV protocol. Two-person interactions usually require the movement of the whole body. Top layers of our tailored network architecture can learn the whole body motion better than the naive CNN models originally designed for detecting generic objects in a still image. 

On the other hand, it appears that our method performs poorly on two classes, namely ``brushing teeth'' (58.3\%) and ``brushing hair'' (47.6\%). Confusion matrix reveals that ``brushing teeth'' is often misclassified as either ``cheer up'' and ``hand waving'', while the ``brushing hair'' is misclassified as ``hand waving''. This may be because the ``head joint'', which is selected as the reference joint for the torso, is not stationary enough compared to the other reference joints in these action types. 

\textbf{SBU Kinect Interaction Dataset}
Table \ref{tab:table2} shows the comparisons of our proposed method with the previous studies on SBU dataset. As can be seen, our proposed method achieved the best performance on this dataset over all the other previous methods. In particular, our method gains more than 5.0 points improvement compared to the two state-of-the-art CNN-based methods \cite{ke2017skeletonnet, ke2017new}, about 4.0 points better than \cite{liu2018skeleton}, and approximately 2.0 points better than \cite{zhang2017view}. These results again confirm that our method has superior performance on two-person interaction actions.
%\iffalse
%\fi

\iffalse
\begin{table}[t!]
	\caption{Classification Performance on SBU Dataset}
	\label{tab:table2}
	\begin{tabular}{c|c}
		\hline
		Methods& Acc.\\
		\hline
		Deep LSTM+Co-occurence \cite{zhu2016co} & 90.4\\
		ST-LSTM+Trust Gate \cite{liu2016spatio} & 93.3\\
		SkeletonNet \cite{ke2017skeletonnet} & 93.5 \\
		Clips+CNN+Concatenation \cite{ke2017new} & 92.9 \\
		Clips+CNN+MTLN \cite{ke2017new} & 93.6 \\
		Context-aware  attention LSTM \cite{liu2018skeleton} & 94.9 \\
		\hline
		\textbf{F2CSkeleton (Proposed)} & \textbf{99.1}\\ 
	\end{tabular}
\end{table}
\fi
\iffalse
\begin{table}[t!]
	\centering %\begin{center}
	\small
	\caption{Classification Performance with Two-person Interactions}
	\label{tab:table2}
	\begin{tabular}{c|*{2}c}
		\hline
		Actions& \multicolumn{1}{c}{Cross-view}\\
		{} & SkeletonNet & Proposed method \\
			\hline
		Punching/slapping & 57.6 & \textbf{82.1}\\
		Kicking& 54.4 & \textbf{88.8} \\
		Pushing & 70.9 & \textbf{86.6} \\
		Pat on back& 50.1 & \textbf{78.9} \\
		Point finger& 53.9 & \textbf{90.0} \\
		Hugging& 80.5 & \textbf{86.8} \\
		Giving something & 72.5 & \textbf{90.4} \\
		Touch other's pocket & 57.7 & \textbf{91.7} \\
		Handshaking& 82.9 & \textbf{95.8} \\
		Walking towards& 73.3 & \textbf{98.0} \\
		Walking apart& 69.2 & \textbf{71.5} \\
	\end{tabular}
	%\end{center}
\end{table}
\fi

\section{Conclusion}
	\label{sec:conclusion}
This paper addresses two problems of the previous studies: the loss of temporal information in a skeleton sequence when modeling using CNNs and the need for a network model specific to a human skeleton sequence. We first propose to segment a skeleton sequence to retrieve the dependencies between temporal segments in an action. We also propose an F2C CNN architecture for exploiting the spatio-temporal feature of skeleton data. As a result, our method with only three network blocks shows the superior generalization ability across very deep CNN models. We achieve a performance of 79.6\% and 84.6\% of accuracies on the large skeleton dataset, NTU RGB+D,  with cross-object and cross-view protocol, respectively, which reaches the state-of-the-art.   

In the future, as has been noted, we will adopt the notion of multi-task learning. In addition, since we do not share weights between input slices during training, our network has more trainable parameters compared to general CNN models with the same input size and the number of filters. We believe our method will work better if we reduce the number of feature maps in convolutional layers. The current skeleton data is very challenging due to noisy joints. For example, by manually checking skeleton data from the first data collection setup of NTU RGB+D,  we find that there were about 8.8\% of noisy detections. Because our method did not apply any algorithms to remove these noises from the input, it is promising to take this into consideration for better performance.
    
\subsubsection*{Acknowledgments}
This work was supported by JSPS KAKENHI 15K12061 and by JST CREST Grant Number JPMJCR1687, Japan.

\bibliographystyle{apalike}
\bibliography{arxiv}

\begin{thebibliography}{36}
\providecommand{\natexlab}[1]{#1}
\providecommand{\url}[1]{\texttt{#1}}
\expandafter\ifx\csname urlstyle\endcsname\relax
  \providecommand{\doi}[1]{doi: #1}\else
  \providecommand{\doi}{doi: \begingroup \urlstyle{rm}\Url}\fi

\bibitem[Chatfield et~al.(2014)Chatfield, Simonyan, Vedaldi, and
  Zisserman]{Chatfield14}
K.~Chatfield, K.~Simonyan, A.~Vedaldi, and A.~Zisserman.
\newblock Return of the devil in the details: Delving deep into convolutional
  nets.
\newblock In \emph{British Machine Vision Conference (BMVC)}, 2014.

\bibitem[Du et~al.(2015)Du, Wang, and Wang]{du2015hierarchical}
Yong Du, Wei Wang, and Liang Wang.
\newblock Hierarchical recurrent neural network for skeleton based action
  recognition.
\newblock In \emph{Proc. of Computer Vision and Pattern Recognition (CVPR)},
  pages 1110--1118, 2015.

\bibitem[George et~al.(2017)George, Lehrach, Kansky, L{\'a}zaro-Gredilla, Laan,
  Marthi, Lou, Meng, Liu, Wang, et~al.]{george2017generative}
Dileep George, Wolfgang Lehrach, Ken Kansky, Miguel L{\'a}zaro-Gredilla,
  Christopher Laan, Bhaskara Marthi, Xinghua Lou, Zhaoshi Meng, Yi~Liu, Huayan
  Wang, et~al.
\newblock A generative vision model that trains with high data efficiency and
  breaks text-based captchas.
\newblock \emph{Science}, 358\penalty0 (6368):\penalty0 eaag2612, 2017.

\bibitem[Han et~al.(2017)Han, Reily, Hoff, and Zhang]{han2017space}
Fei Han, Brian Reily, William Hoff, and Hao Zhang.
\newblock Space-time representation of people based on 3d skeletal data: A
  review.
\newblock \emph{Proc. of Computer Vision and Image Understanding},
  158:\penalty0 85--105, 2017.

\bibitem[He et~al.(2016)He, Zhang, Ren, and Sun]{he2016deep}
Kaiming He, Xiangyu Zhang, Shaoqing Ren, and Jian Sun.
\newblock Deep residual learning for image recognition.
\newblock In \emph{Proc. of Computer Vision and Pattern Recognition (CVPR)},
  pages 770--778, 2016.

\bibitem[Hou and Andrews(1978)]{hou1978cubic}
Hsieh Hou and H~Andrews.
\newblock Cubic splines for image interpolation and digital filtering.
\newblock \emph{IEEE Transactions on acoustics, speech, and signal processing},
  26\penalty0 (6):\penalty0 508--517, 1978.

\bibitem[Hu et~al.(2017)Hu, Liu, Li, Song, and Liu]{hutemporal}
Yueyu Hu, Chunhui Liu, Yanghao Li, Sijie Song, and Jiaying Liu.
\newblock Temporal perceptive network for skeleton-based action recognition.
\newblock In \emph{Proc. of British Machine Vision Conference (BMVC)}, pages
  1--2, 2017.

\bibitem[Ke et~al.(2017{\natexlab{a}})Ke, An, Bennamoun, Sohel, and
  Boussaid]{ke2017skeletonnet}
Qiuhong Ke, Senjian An, Mohammed Bennamoun, Ferdous Sohel, and Farid Boussaid.
\newblock Skeletonnet: Mining deep part features for 3-d action recognition.
\newblock \emph{IEEE signal processing letters}, 24\penalty0 (6):\penalty0
  731--735, 2017{\natexlab{a}}.

\bibitem[Ke et~al.(2017{\natexlab{b}})Ke, Bennamoun, An, Sohel, and
  Boussaid]{ke2017new}
Qiuhong Ke, Mohammed Bennamoun, Senjian An, Ferdous Sohel, and Farid Boussaid.
\newblock A new representation of skeleton sequences for 3d action recognition.
\newblock In \emph{Proc. of Computer Vision and Pattern Recognition (CVPR)},
  pages 4570--4579. IEEE, 2017{\natexlab{b}}.

\bibitem[Kerola et~al.(2016)Kerola, Inoue, and Shinoda]{kerola2016graph}
Tommi Kerola, Nakamasa Inoue, and Koichi Shinoda.
\newblock Graph regularized implicit pose for 3d human action recognition.
\newblock In \emph{Proc. of Asia-Pacific Signal and Information Processing
  Association Annual Summit and Conference (APSIPA)}, pages 1--4. IEEE, 2016.

\bibitem[Kim and Reiter(2017)]{kim2017interpretable}
Tae~Soo Kim and Austin Reiter.
\newblock Interpretable 3d human action analysis with temporal convolutional
  networks.
\newblock In \emph{Computer Vision and Pattern Recognition Workshops (CVPRW)},
  pages 1623--1631. IEEE, 2017.

\bibitem[Kingma and Ba(2015)]{kingma2014adam}
Diederik~P Kingma and Jimmy Ba.
\newblock Adam: A method for stochastic optimization.
\newblock \emph{Proc. of International Conference on Learning Representations
  (ICLR)}, 2015.

\bibitem[Krizhevsky et~al.(2012)Krizhevsky, Sutskever, and
  Hinton]{krizhevsky2012imagenet}
Alex Krizhevsky, Ilya Sutskever, and Geoffrey~E Hinton.
\newblock Imagenet classification with deep convolutional neural networks.
\newblock In \emph{Proc. of Advances in Neural Information Processing Systems
  (NIPS)}, pages 1097--1105, 2012.

\bibitem[Liu et~al.(2016)Liu, Shahroudy, Xu, and Wang]{liu2016spatio}
Jun Liu, Amir Shahroudy, Dong Xu, and Gang Wang.
\newblock Spatio-temporal lstm with trust gates for 3d human action
  recognition.
\newblock In \emph{Proc. of European Conference on Computer Vision}, pages
  816--833. Springer, 2016.

\bibitem[Liu et~al.(2017{\natexlab{a}})Liu, Shahroudy, Xu, Chichung, and
  Wang]{liu2017skeleton}
Jun Liu, Amir Shahroudy, Dong Xu, Alex~Kot Chichung, and Gang Wang.
\newblock Skeleton-based action recognition using spatio-temporal lstm network
  with trust gates.
\newblock \emph{IEEE Transactions on Pattern Analysis and Machine
  Intelligence}, 2017{\natexlab{a}}.

\bibitem[Liu et~al.(2018)Liu, Wang, Duan, Abdiyeva, and Kot]{liu2018skeleton}
Jun Liu, Gang Wang, Ling-Yu Duan, Kamila Abdiyeva, and Alex~C Kot.
\newblock Skeleton-based human action recognition with global context-aware
  attention lstm networks.
\newblock \emph{IEEE Transactions on Image Processing}, 27\penalty0
  (4):\penalty0 1586--1599, 2018.

\bibitem[Liu et~al.(2017{\natexlab{b}})Liu, Chen, and Liu]{liu20173d}
Mengyuan Liu, Chen Chen, and Hong Liu.
\newblock 3d action recognition using data visualization and convolutional
  neural networks.
\newblock In \emph{IEEE International Conference on Multimedia and Expo
  (ICME)}, pages 925--930. IEEE, 2017{\natexlab{b}}.

\bibitem[Liu et~al.(2017{\natexlab{c}})Liu, Liu, and Chen]{liu2017enhanced}
Mengyuan Liu, Hong Liu, and Chen Chen.
\newblock Enhanced skeleton visualization for view invariant human action
  recognition.
\newblock \emph{Pattern Recognition}, 68:\penalty0 346--362,
  2017{\natexlab{c}}.

\bibitem[Long et~al.(2015)Long, Cao, Wang, and Jordan]{long2015learning}
Mingsheng Long, Yue Cao, Jianmin Wang, and Michael~I Jordan.
\newblock Learning transferable features with deep adaptation networks.
\newblock In \emph{Proc. of the 32nd International Conference on Machine
  Learning, {ICML}}, pages 97--105, 2015.

\bibitem[Luo et~al.(2013)Luo, Wang, and Qi]{luo2013group}
Jiajia Luo, Wei Wang, and Hairong Qi.
\newblock Group sparsity and geometry constrained dictionary learning for
  action recognition from depth maps.
\newblock In \emph{Proc. of Computer vision (ICCV)}, pages 1809--1816. IEEE,
  2013.

\bibitem[Pan and Yang(2010)]{pan2010survey}
Sinno~Jialin Pan and Qiang Yang.
\newblock A survey on transfer learning.
\newblock \emph{IEEE Transactions on knowledge and data engineering},
  22\penalty0 (10):\penalty0 1345--1359, 2010.

\bibitem[Presti and La~Cascia(2016)]{presti20163d}
Liliana~Lo Presti and Marco La~Cascia.
\newblock 3d skeleton-based human action classification: A survey.
\newblock \emph{Proc. of Pattern Recognition}, 53:\penalty0 130--147, 2016.

\bibitem[Russakovsky et~al.(2015)Russakovsky, Deng, Su, Krause, Satheesh, Ma,
  Huang, Karpathy, Khosla, Bernstein, Berg, and Fei-Fei]{ILSVRC15}
Olga Russakovsky, Jia Deng, Hao Su, Jonathan Krause, Sanjeev Satheesh, Sean Ma,
  Zhiheng Huang, Andrej Karpathy, Aditya Khosla, Michael Bernstein,
  Alexander~C. Berg, and Li~Fei-Fei.
\newblock {ImageNet Large Scale Visual Recognition Challenge}.
\newblock \emph{International Journal of Computer Vision (IJCV)}, 115\penalty0
  (3):\penalty0 211--252, 2015.
\newblock \doi{10.1007/s11263-015-0816-y}.

\bibitem[Shahroudy et~al.(2016)Shahroudy, Liu, Ng, and Wang]{shahroudy2016ntu}
Amir Shahroudy, Jun Liu, Tian-Tsong Ng, and Gang Wang.
\newblock Ntu rgb+d: A large scale dataset for 3d human activity analysis.
\newblock In \emph{Proc. of Computer Vision and Pattern Recognition (CVPR)},
  June 2016.

\bibitem[Simonyan and Zisserman(2014)]{simonyan2014very}
Karen Simonyan and Andrew Zisserman.
\newblock Very deep convolutional networks for large-scale image recognition.
\newblock \emph{Proc. of CoRR}, 2014.

\bibitem[Song et~al.(2017)Song, Lan, Xing, Zeng, and Liu]{song2017end}
Sijie Song, Cuiling Lan, Junliang Xing, Wenjun Zeng, and Jiaying Liu.
\newblock An end-to-end spatio-temporal attention model for human action
  recognition from skeleton data.
\newblock In \emph{Proc. of Association for the Advancement of Artificial
  Intelligence (AAAI)}, volume~1, page~7, 2017.

\bibitem[Vemulapalli et~al.(2014)Vemulapalli, Arrate, and
  Chellappa]{vemulapalli2014human}
Raviteja Vemulapalli, Felipe Arrate, and Rama Chellappa.
\newblock Human action recognition by representing 3d skeletons as points in a
  lie group.
\newblock In \emph{Proc. of Computer Vision and Pattern Recognition}, pages
  588--595, 2014.

\bibitem[Wagner et~al.(2013)Wagner, Thom, Schweiger, Palm, and
  Rothermel]{wagner2013learning}
Raimar Wagner, Markus Thom, Roland Schweiger, Gunther Palm, and Albrecht
  Rothermel.
\newblock Learning convolutional neural networks from few samples.
\newblock In \emph{Neural Networks (IJCNN), The 2013 International Joint
  Conference on}, pages 1--7. IEEE, 2013.

\bibitem[Wang et~al.(2012)Wang, Liu, Wu, and Yuan]{wang2012mining}
Jiang Wang, Zicheng Liu, Ying Wu, and Junsong Yuan.
\newblock Mining actionlet ensemble for action recognition with depth cameras.
\newblock In \emph{Proc. of Computer Vision and Pattern Recognition (CVPR)},
  pages 1290--1297. IEEE, 2012.

\bibitem[Yang and Tian(2014)]{yang2014effective}
Xiaodong Yang and YingLi Tian.
\newblock Effective 3d action recognition using eigenjoints.
\newblock \emph{Journal of Visual Communication and Image Representation},
  25\penalty0 (1):\penalty0 2--11, 2014.

\bibitem[Yun et~al.(2012)Yun, Honorio, Chattopadhyay, Berg, and
  Samaras]{yun2012two}
Kiwon Yun, Jean Honorio, Debaleena Chattopadhyay, Tamara~L Berg, and Dimitris
  Samaras.
\newblock Two-person interaction detection using body-pose features and
  multiple instance learning.
\newblock In \emph{IEEE Computer Society Conference on Computer Vision and
  Pattern Recognition Workshops (CVPRW)}, pages 28--35. IEEE, 2012.

\bibitem[Zanfir et~al.(2013)Zanfir, Leordeanu, and
  Sminchisescu]{zanfir2013moving}
Mihai Zanfir, Marius Leordeanu, and Cristian Sminchisescu.
\newblock The moving pose: An efficient 3d kinematics descriptor for
  low-latency action recognition and detection.
\newblock In \emph{Proc. of the IEEE International Conference on Computer
  Vision}, pages 2752--2759, 2013.

\bibitem[Zhang et~al.(2017{\natexlab{a}})Zhang, Lan, Xing, Zeng, Xue, and
  Zheng]{zhang2017view}
Pengfei Zhang, Cuiling Lan, Junliang Xing, Wenjun Zeng, Jianru Xue, and Nanning
  Zheng.
\newblock View adaptive recurrent neural networks for high performance human
  action recognition from skeleton data.
\newblock \emph{International Conference on Computer Vision}, pages 2136--2145,
  2017{\natexlab{a}}.

\bibitem[Zhang et~al.(2017{\natexlab{b}})Zhang, Liu, and
  Xiao]{zhang2017geometric}
Songyang Zhang, Xiaoming Liu, and Jun Xiao.
\newblock On geometric features for skeleton-based action recognition using
  multilayer lstm networks.
\newblock In \emph{IEEE Winter Conference on Applications of Computer Vision
  (WACV)}, pages 148--157. IEEE, 2017{\natexlab{b}}.

\bibitem[Zhang and Yeung(2014)]{zhang2014regularization}
Yu~Zhang and Dit-Yan Yeung.
\newblock A regularization approach to learning task relationships in multitask
  learning.
\newblock \emph{ACM Transactions on Knowledge Discovery from Data (TKDD)},
  8\penalty0 (3):\penalty0 12, 2014.

\bibitem[Zhu et~al.(2016)Zhu, Lan, Xing, Zeng, Li, Shen, Xie,
  et~al.]{zhu2016co}
Wentao Zhu, Cuiling Lan, Junliang Xing, Wenjun Zeng, Yanghao Li, Li~Shen,
  Xiaohui Xie, et~al.
\newblock Co-occurrence feature learning for skeleton based action recognition
  using regularized deep lstm networks.
\newblock In \emph{Proc. of Association for the Advancement of Artificial
  Intelligence (AAAI)}, volume~2, page~8, 2016.

\end{thebibliography}

\end{document}